%% file: acl_latex.tex
\pdfoutput=1

\documentclass[11pt]{article}

\usepackage[final]{acl}

\usepackage{times}
\usepackage{latexsym}

\usepackage[T1]{fontenc}

\usepackage[utf8]{inputenc}

\usepackage{microtype}

\usepackage{inconsolata}

\usepackage{graphicx}

\usepackage{multirow}
\usepackage{fixmath,mathtools,nicefrac}
\usepackage{url}            
\usepackage{booktabs}       
\usepackage{amsfonts}       
\usepackage{nicefrac}       
\usepackage{microtype}      
\usepackage{makecell}
\usepackage{adjustbox}
\usepackage{pifont}
\usepackage{bbding}
\usepackage{enumerate}
\usepackage{enumitem}
\usepackage{subcaption}
\usepackage{comment}
\usepackage{caption}
\usepackage{float}
\usepackage{wrapfig} 
\usepackage[misc]{ifsym} 
\definecolor{mygray1}{gray}{.95}
\definecolor{mygray2}{gray}{.9}
\definecolor{mygray3}{gray}{.95}
\usepackage{pifont}

\usepackage{arydshln}

\usepackage[capitalize]{cleveref}
\crefname{section}{Sec.}{Secs.}
\Crefname{section}{Section}{Sections}
\Crefname{table}{Table}{Tables}
\crefname{table}{Tab.}{Tabs.}
\Crefname{figure}{Figure}{Figures}
\crefname{figure}{Fig.}{Figs.}
\crefname{appendix}{Appx.}{Appxs.}
\usepackage{tcolorbox}

\tcbuselibrary{most}
\tcbuselibrary{skins}
\tcbset{
  aibox/.style={
    top=10pt,
    colback=white,
    colframe=black,
    colbacktitle=black,
    enhanced,
    center,
    attach boxed title to top left={yshift=-0.1in,xshift=0.15in},
    boxed title style={boxrule=0pt,colframe=white,},
  }
}
\newtcolorbox{AIbox}[2][]{aibox, title=#2,#1}
\newcommand{\tablestyle}[2]{\setlength{\tabcolsep}{#1}\renewcommand{\arraystretch}{#2}\centering\footnotesize}
\newlength\savewidth\newcommand\shline{\noalign{\global\savewidth\arrayrulewidth
		\global\arrayrulewidth .8pt}\hline\noalign{\global\arrayrulewidth\savewidth}}		

\title{OmniAlign-V: Towards Enhanced Alignment \\  of MLLMs with Human Preference}

\author{Xiangyu Zhao$^{1,2}$\footnotemark[1] \quad  Shengyuan Ding$^{2,3}$\footnotemark[1] \quad Zicheng Zhang$^{1,2}$ \\ \bfseries Haian Huang$^2$ \quad Maosong Cao$^2$ \quad Weiyun Wang$^{2,4}$ \quad Jiaqi Wang$^2$ \quad Xinyu Fang$^{2,5}$ \\ \bfseries Wenhai Wang$^2$ \quad Guangtao Zhai$^{1,2}$ \quad Haodong Duan$^2$\footnotemark[2] \quad Hua Yang$^1$\footnotemark[2] \quad Kai Chen$^2$\footnotemark[2] \\ Shanghai Jiaotong University$^1$ \quad Shanghai AI Laboratory$^2$ \quad Nanjing University$^3$ \\ Fudan University$^4$  \quad Zhejiang University$^5$}

\begin{document}
\twocolumn[{%
\renewcommand\twocolumn[1][]{#1}%
\maketitle
\vspace{-20mm}

\begin{center}
    \centering
    \captionsetup{type=figure}
    \vspace{10mm}
    \includegraphics[width=\linewidth]{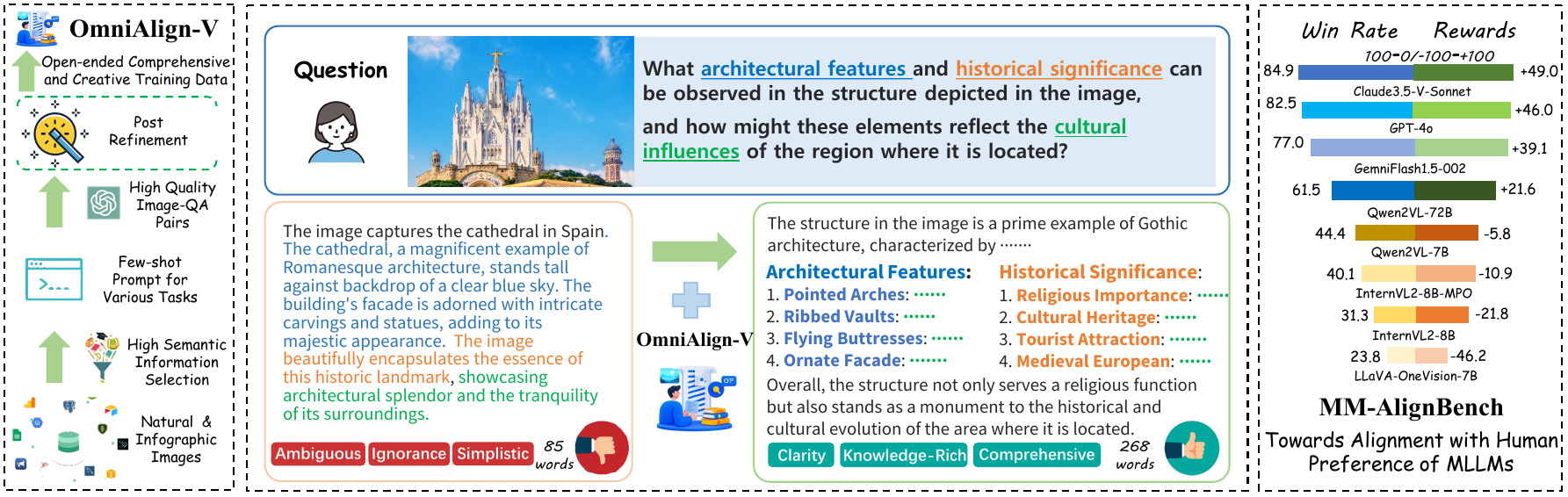}
    \caption{\textbf{OmniAlign-V} consists of curated images paired with open-ended, comprehensive question-answer pairs, significantly improving the alignment of MLLMs with human preferences. 
    Additionally, we introduce \textbf{MM-AlignBench}, a human-annotated, high-quality benchmark specifically designed to evaluate the ability of MLLMs to align with human values.}
    \label{fig:spotlight}
    \vspace{3mm}
\end{center}%
}
]
\footnotetext[1]{Equal Contribution.}
\footnotetext[2]{Corresponding Author.}
\input{latex/0_abs}
\input{latex/1_intro}
\input{latex/2_related}

\input{latex/3_subjective_cap}
\input{latex/4_data_engine}

\input{latex/5_main_results}
\bibliography{custom}

\input{latex/6_appd}


\end{document}

%% file: latex/0_abs.tex
\begin{abstract}

Recent advancements in open-source multi-modal large language models (MLLMs) have primarily focused on enhancing foundational capabilities, 
leaving a significant gap in human preference alignment. 
This paper introduces \textbf{OmniAlign-V},  a comprehensive dataset of 200K high-quality training samples featuring diverse images, complex questions, and varied response formats to improve MLLMs' alignment with human preferences. 
We also present \textbf{MM-AlignBench}, a human-annotated benchmark specifically designed to evaluate MLLMs' alignment with human values. 
Experimental results show that finetuning MLLMs with OmniAlign-V, 
using Supervised Fine-Tuning (SFT) or Direct Preference Optimization (DPO),
significantly enhances human preference alignment while maintaining or enhancing performance on standard VQA benchmarks, preserving their fundamental capabilities.
Our datasets, benchmark, code and checkpoints have been released at \url{https://github.com/PhoenixZ810/OmniAlign-V}.

\end{abstract}

%% file: latex/1_intro.tex
\section{Introduction}

With the rapid advancement of large language models (LLMs)~\cite{2022chatgpt,touvron2023llama2},  multi-modal large language models (MLLMs)~\cite{OpenAI2023GPT4TR,team2023gemini} have also seen significant improvements. 
Most open-source MLLMs~\cite{liu2023visual,chen2023internvl} are developed by 
connecting a vision encoder~\cite{dosovitskiy2020image} to a pretrained LLM,
followed by vision instruction tuning.
Existing vision instruction tuning datasets~\cite{liu2023visual,chen2023sharegpt4v,chen2024allava} and 
multi-modal evaluation benchmarks~\cite{liu2023mmbench,yue2023mmmu,lu2023mathvista} primarily focus on assessing fundamental capabilities (object recognition, OCR, \emph{etc.}) of MLLMs,
while paying little attention to human preference alignment.
Consequently, 
while open-source MLLMs achieve comparable performance to proprietary counterparts on objective metrics of these foundational skills,
they display a significant gap in the alignment with human preferences, 
which detrimentally impacts user experience in multi-modal conversational interactions, as demonstrated in Appx.~\ref{appd: align}.
%

In this work, 
a preliminary investigation was conducted to quantitatively assess the degradation of human preference alignment in MLLMs(\cref{sec: preliminary}). 
Experimental results on text-only subjective evaluation benchmarks~\cite{dubois2024length,li2024crowdsourced}
revealed that MLLMs exhibited a substantial performance drop compared to their corresponding LLMs. 
A plausible hypothesis suggests that MLLMs suffer from catastrophic forgetting during vision instruction tuning. 
Following this hypothesis, a straightforward solution would be incorporating both multi-modal and text-only SFT data~\cite{xu2024magpie,cao2025condor,ding2023enhancing} for joint learning. 
However, our experiments (Tab.~\ref{tab: mm}) indicate that this approach not only failed to yield improvements
in human preference alignment under multi-modal contexts but also demonstrated negative impacts on the foundational multi-modal skills of MLLMs. 
These observations suggest that enhancing human alignment in multi-modal scenarios necessitates the development of specialized multi-modal instruction tuning datasets.

Existing multi-modal instruction tuning datasets predominantly focus on foundational capabilities, featuring simple language patterns and uniform response styles (\cref{fig:badsft} in appendix).
This study argues that effective multi-modal datasets for enhancing human preference alignment should incorporate several critical characteristics: \textbf{open-ended questions}, \textbf{broad topic coverage}, \textbf{diverse response formats} (varying in length and style), and \textbf{strict adherence to instructions}. 
Based on these principles, we constructed \textbf{OmniAlign-V}.
In terms of image sources, OmniAlign-V encompasses natural images and infographics such as posters and charts. 
Furthermore, a novel solution was developed to filter out semantically rich images from natural image collections.
With respect to tasks, complex knowledge-based question answering, creative tasks, and reasoning tasks were designed for different image types. 
Each task category incorporates diverse subtasks, and state-of-the-art MLLMs were leveraged to obtain diverse, high-quality responses. 
Ultimately, OmniAlign-V comprises $\sim$200K multi-modal instruction tuning samples, 
exhibiting a significantly different overall data distribution compared to existing multi-modal SFT datasets. 

Leveraging OmniAlign-V, comprehensive experiments were conducted to explore its full potential for enhancing human alignment in MLLMs. 
Integrating OmniAlign-V into the SFT stage of the LLaVA-NeXT~\cite{liu2024llavanext} structure with InternLM2.5-7B and Qwen2.5-32B yielded significant improvements in human preference alignment across various powerful LLM encoders, 
including InternLM2.5~\cite{cai2024internlm2} and Qwen2.5~\cite{qwen}. 
Furthermore, on ground-truth-based VQA benchmarks like MMMU~\cite{yue2023mmmu} or OCRBench~\cite{liu2023hidden}, 
MLLMs finetuned with OmniAlign-V displayed comparable or superior performance. 
Beyond its application in SFT, 
OmniAlign-V also demonstrated additional value when applied to Direct Preference Optimization (DPO)~\cite{rafailov2024direct}. 
Experimental results indicate that implementing OmniAlign-V for DPO yields further improvements in preference alignment, 
surpassing baseline models finetuned on the same dataset. After the SFT and DPO stage, \textbf{our LLaVA-Next baseline with Qwen2.5-32B surpasses the state-of-the-art model Qwen2VL-72B}~\cite{chen2024far} finetuned on extensive proprietary datasets.

Throughout the exploration, 
we also observed that existing multi-modal human preference benchmarks~\cite{lu2024wildvision,qian2024mia} lack diversity in image sources, contain repetitive questions, and lack clarity. 
To address these limitations, 
we introduce \textbf{MM-AlignBench},
a high-quality benchmark comprising 252 carefully curated samples with diverse image sources and meticulously crafted questions by human annotators, enabling comprehensive evaluation of MLLMs' alignment with human preferences.
MM-AlignBench, along with other MLLM human preference benchmarks, 
was employed throughout this study for evaluation.




The contributions of this study are as follows:

1. An in-depth investigation into the degradation of MLLM human alignment, 
analyzing the impact of both text-only and multi-modal tuning data.

2. The introduction of OmniAlign-V, 
a comprehensive open-ended multi-modal SFT dataset, complemented by OmniAlign-V-DPO. Extensive experiments demonstrate the effectiveness of these datasets in improving human preference alignment.

3. The development of MM-AlignBench, a carefully curated benchmark comprising high-quality, human-annotated samples specifically designed to evaluate MLLMs' human preference alignment.

%% file: latex/2_related.tex
\section{Related Work}

\textbf{Alignment of LLMs.} Alignment, encompassing the ability to follow human instructions and provide helpful assistance~\cite{liu2024trustworthyllmssurveyguideline}, has long been a critical aspect of LLMs. 
Recent works~\cite{wang2023openchat,xu2024magpie,cao2025condor} focused on generating high-quality SFT training data to enhance the alignment of LLMs. 
Besides, recent benchmarks~\cite{liu2023alignbench,alpaca_eval,li2024crowdsourced} have introduced open-ended and challenging questions to assess the alignment performance of LLMs. 
However, there is a notable lack of benchmarks designed to evaluate the alignment of MLLMs.

\noindent\textbf{Visual Question Answering Datasets.} In the early stages, Visual Question Answering (VQA) datasets were primarily used for basic semantic alignment between images and text~\cite{radford2021learning,li2022blip, zhao2024open}. These datasets were relatively simple in structure. Popular VQA datasets~\cite{hudson2019gqa, antol2015vqa} predominantly featured straightforward questions that elicited single-word or short-sentence answers. Similarly, visual classification and detection datasets~\cite{lin2014microsoft, schuhmann2021laion, sharma2018conceptual} typically consisted of brief descriptions that focused on the main object within the image.

\noindent\textbf{Instruction Tuning data.} With the rapid development of MLLMs~\cite{chen2024far, bai2023qwen}, the instruction tuning data has gained more attention. LLaVA~\cite{liu2023visual} leveraged advanced LLMs to generate instruction-following formatted data from traditional VQA datasets. Recent works have further expanded this approach by employing MLLM to generate captions~\cite{chen2023sharegpt4v}, complex question-answer pairs~\cite{ chen2024allava, gu2024infinity}, OCR data~\cite{textocr-gpt4v} and math-related data~\cite{shi2024math}. Other efforts~\cite{tong2024cambrian, li2024llava} aggregates multiple publicly available document image datasets into unified resources. However, these datasets primarily focus on fundamental visual capabilities with short dialogues and factual questions, resulting in suboptimal alignment with human preferences.

%% file: latex/3_subjective_cap.tex
\section{Human Alignment of MLLMs: The Preliminary Study}
\label{sec: preliminary}
When handling open-ended and flexible questions involving images,
state-of-the-art open-source MLLMs -- 
despite excelling in certain recognition tasks -- 
exhibit significantly weaker alignment with human preferences compared to GPT-4o (\cref{fig:internvl} in appendix).
We hypothesize that this decline in multi-modal preference alignment stems from a reduction in the language model's proficiency after the multi-modal SFT stage. 
To test this hypothesis, we evaluated state-of-the-art MLLMs on text-only human preference alignment benchmarks~\cite{liu2023alignbench,alpaca_eval,li2024crowdsourced}.
Additionally, we constructed a LLaVA~\cite{liu2023visual} baseline using InternLM-2.5-7B, fine-tuned on the LLaVA-Next-778k~\cite{liu2023improved} SFT dataset. As shown in \cref{tab: subjective language}, 
the LLM’s ability to handle text-only open-ended questions degrades significantly after multi-modal SFT. 
This degradation may be due to (1) insufficient quantity or quality of text-only samples during multi-modal SFT, or (2) the overly simplistic style of multi-modal fine-tuning data derived from traditional VQA datasets~\cite{vqav2,hudson2019gqa}, often consisting of simple questions and short, factual answers.

\begin{table}
    \centering
    \resizebox{.5\textwidth}{!}{%
    \begin{tabular}{ll|ccc}
    \Xhline{0.15em}
        \textbf{Method} & \textbf{Type} & \textbf{AlignBench} & \textbf{AlpacaEval-V2} & \textbf{ArenaHard}  \\
        \hline
        InternLM2.5-7B & LLM  &6.36& 27.58& 27.06 \\
        InternVL2-8B   & MLLM &4.04(-36.4\%)& 3.35(-87.9\%)& 4.65(-82.8\%) \\
        LLaVA-Internlm & MLLM &4.66(-26.7\%)& 4.22(-84.7\%)& 4.93(-81.8\%) \\
        \hline
        Qwen2-7B     & LLM &6.02& 24.47& 32.84 \\
        Qwen2VL-7B & MLLM &4.92(-18.3\%)& 3.85(-83.4\%)& 6.46(-80.3\%) \\
        \hline
        LLaMA3-8B     & LLM &4.88& 30.19& 31.96 \\
        MiniCPM-V2.5 & MLLM &3.84(-21.3\%)& 7.33(-75.7\%)& 8.05(-74.8\%) \\
        \hline
        InternLM2-20B & LLM  &5.49& 43.35& 33.75 \\
        InternVL2-26B & MLLM &4.39(-20.0\%)& 5.34(-87.7\%)& 11.25(-66.7\%) \\
        \hline
        Hermes2-llama3-70b & LLM &5.72 &46.09 &57.40 \\
        InternVL2-76B     & MLLM &4.33(-24.3\%) &8.32(-81.9\%) &16.17(-72.3\%) \\
    \Xhline{0.15em}
    \end{tabular}
    }%
    \caption{\textbf{Language Alignment Benchmark Results.} After multi-modal SFT, MLLMs demonstrate significant decline in human preference alignment compared to their corresponding LLMs.} 
    \label{tab: subjective language}
    \vspace{-10pt}
\end{table}

\begin{table*}[t]
    \centering
    \resizebox{\textwidth}{!}{%
    \tablestyle{4pt}{1}
    \begin{tabular}{l|ccc|ccccccc}
    \Xhline{0.15em}
        & \multicolumn{3}{c|}{\textit{\textbf{Language Benchmarks}}} & \multicolumn{5}{c}{\textit{\textbf{Multi-modal Benchmarks}}}  \\
        \textbf{Model}  & \textbf{AlignBench} & \textbf{AlpacaEval-V2} & \textbf{ArenaHard} &\textbf{WildVision} &\textbf{MMVet} &\textbf{MMBench-V1.1}  &\textbf{AI2D} &\textbf{OCRBench} \\
        \hline
        LLaVA$^I$-LLaVANext778k         &4.7& 31.7& 21.6 & 18.4/-55.1& 41.2 &73.7 &74.2 &39.7\\
        LLaVA$^I$-LLaVANext$_{mm}$738k-Magpie40k  &4.5 \color{red}{$\boldsymbol{\downarrow}$}& 65.5 \color{SeaGreen}{$\boldsymbol{\uparrow}$}& 44.6 \color{SeaGreen}{$\boldsymbol{\uparrow}$} & 16.8/-58.9 \color{red}{$\boldsymbol{\downarrow}$}& 37.7 \color{red}{$\boldsymbol{\downarrow}$} &73.1 \color{red}{$\boldsymbol{\downarrow}$} &73.4 \color{red}{$\boldsymbol{\downarrow}$} &38.7 \color{red}{$\boldsymbol{\downarrow}$}\\
        LLaVA$^I$-LLaVANext$_{mm}$738k-Condor40k  &5.6 \color{SeaGreen}{$\boldsymbol{\uparrow}$}& 72.9 \color{SeaGreen}{$\boldsymbol{\uparrow}$}& 55.7 \color{SeaGreen}{$\boldsymbol{\uparrow}$} & 16.8/-57.4 \color{red}{$\boldsymbol{\downarrow}$}& 38.3 \color{red}{$\boldsymbol{\downarrow}$} &72.6 \color{red}{$\boldsymbol{\downarrow}$}  &73.6 \color{red}{$\boldsymbol{\downarrow}$} &38.5 \color{red}{$\boldsymbol{\downarrow}$}\\
        LLaVA$^I$-LLaVANext$_{mm}$738k-Condor80k  &5.7 \color{SeaGreen}{$\boldsymbol{\uparrow}$} & 75.2 \color{SeaGreen}{$\boldsymbol{\uparrow}$} & 55.7 \color{SeaGreen}{$\boldsymbol{\uparrow}$}  & 16.8/-57.0 \color{red}{$\boldsymbol{\downarrow}$}& 38.3 \color{red}{$\boldsymbol{\downarrow}$} &72.6 \color{red}{$\boldsymbol{\downarrow}$}  &74.0 \color{red}{$\boldsymbol{\downarrow}$} &37.6 \color{red}{$\boldsymbol{\downarrow}$}\\
    \Xhline{0.15em}
    \end{tabular}
    }%
    \caption{\textbf{Performance of Incorporating High-Quality Language Data}. LLaVA$^I$ denotes LLaVA structure with InternLM2.5-7B as language model, and LLaVANext$_{mm}$738k refers to the multi-modal data in LLaVANext-778K.
    Integrating high-quality language data significantly improves alignment performance on language benchmarks.
    However, it leads to a decline in multi-modal alignment performance on benchmarks such as WildVision and MMVet. 
    For AlpacaEval and ArenaHard, we present the winning rate against GPT-3.5. }
    \label{tab: mm}
    \vspace{-5pt}
\end{table*}

To assess the impact of the first factor on multi-modal alignment, we examined the role of text-only data in multi-modal SFT.
The LLaVA-Next-778K dataset includes approximately 40K text-only samples from ShareGPT~\cite{vicuna2023}, which are relatively outdated and of lower quality. 
To improve alignment, we experimented with replacing these samples with better-quality alternatives and increasing their proportion in the multi-modal fine-tuning data. 
Specifically, we sampled 40K / 80K instances from two high-quality language SFT datasets: Magpie-Llama3.3~\cite{xu2024magpie} and Condor~\cite{cao2025condor}, using these to replace the original language data in LLaVA-Next. 
We then evaluated models fine-tuned on these different data mixtures using both text-only benchmarks and multi-modal benchmarks, including WildVision~\cite{lu2024wildvision} for human preference alignment and various benchmarks from OpenVLM Leaderboard~\cite{duan2024vlmevalkit} for assessing various fundamental multi-modal capabilities.\footnote{MMVet is open-ended, while others are closed-ended. }


Results in \cref{tab: mm} reveal several key findings. 
While MLLMs tuned with higher-quality text-only samples show significant improvements on text-only alignment benchmarks, 
they unexpectedly demonstrate degraded performance on both multi-modal alignment and common VQA benchmarks. 
This counter-intuitive phenomenon suggests that language alignment capability does not directly translate to multi-modal alignment. 
We therefore argue that high-quality multi-modal human-aligned data is crucial for improving MLLMs' human preference alignment in multi-modal contexts.





%% file: latex/4_data_engine.tex
\section{OmniAlign-V}

Current MLLM instruction tuning datasets primarily focus on enhancing basic capabilities like perception, OCR, and mathematical reasoning. 
These datasets typically contain simple, brief question-answer pairs that inadequately capture human preferences and real-world interaction complexity, as shown in \cref{fig:badsft}. 
We propose that multi-modal training data should also incorporate: (1) \textbf{Open-ended, Diverse, and Creative Questions requiring interdisciplinary knowledge}, (2) \textbf{Comprehensive and Knowledge-rich Responses}. To address these requirements, we develop a novel data synthesis pipeline to generate high-quality human-aligned multi-modal training data, resulting in the creation of \textbf{OmniAlign-V}.


\begin{figure*}[t]
    \centering
    \includegraphics[width=\linewidth]{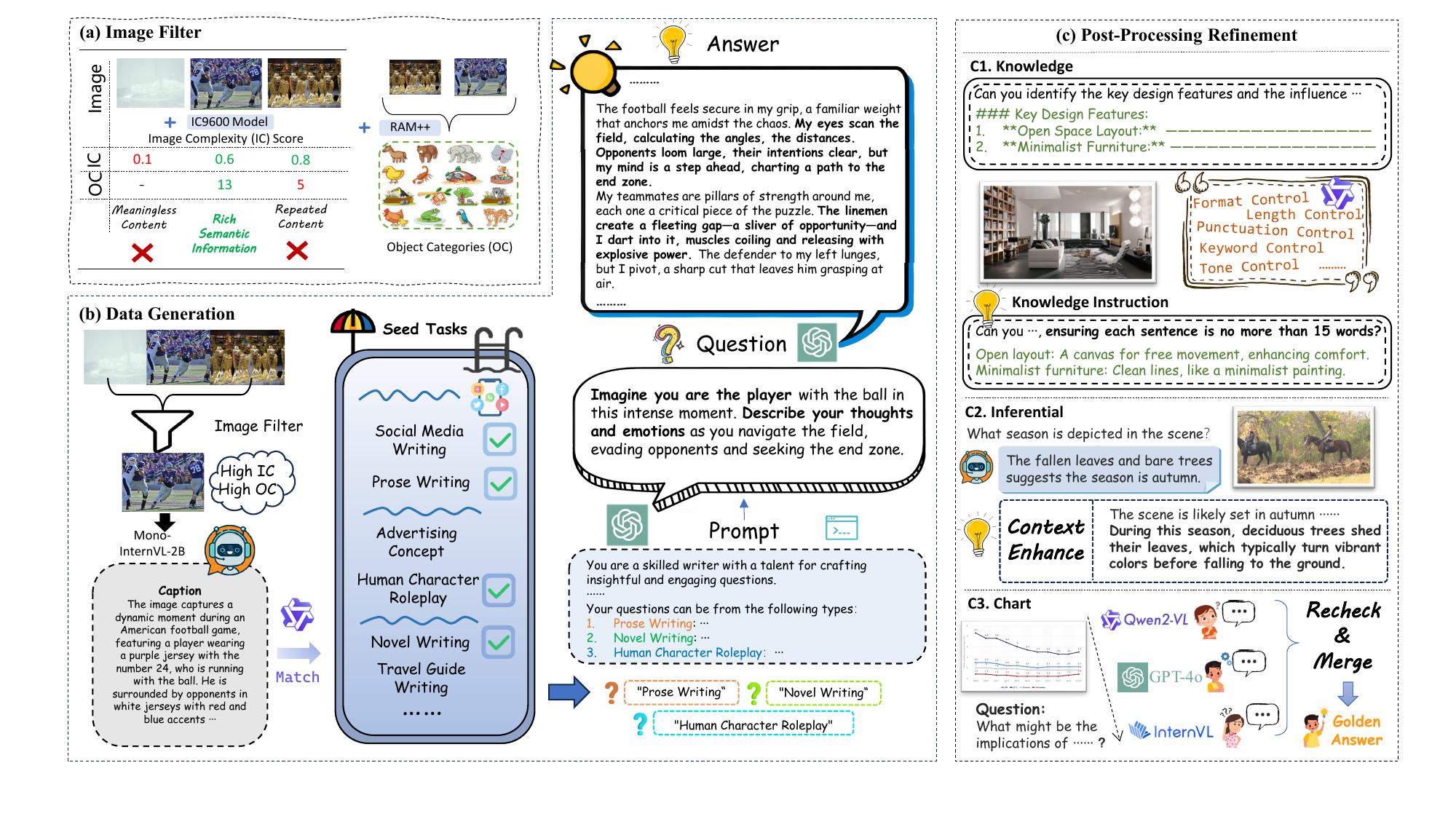}
    \caption{\textbf{Overall pipeline of OmniAlign-V}. By utilizing an image filter and employing a customized pipeline for distinct tasks, we curate semantically rich images paired with high-quality open-ended question-answer sets. Post-refinement further enhances both the variety and quality of our dataset.}
    \label{fig:pipeline}
    \vspace{-10pt}
\end{figure*}

\subsection{Task Taxonomy: A Big Picture}

\begin{figure}[t]
    \centering
    \includegraphics[width=\linewidth]{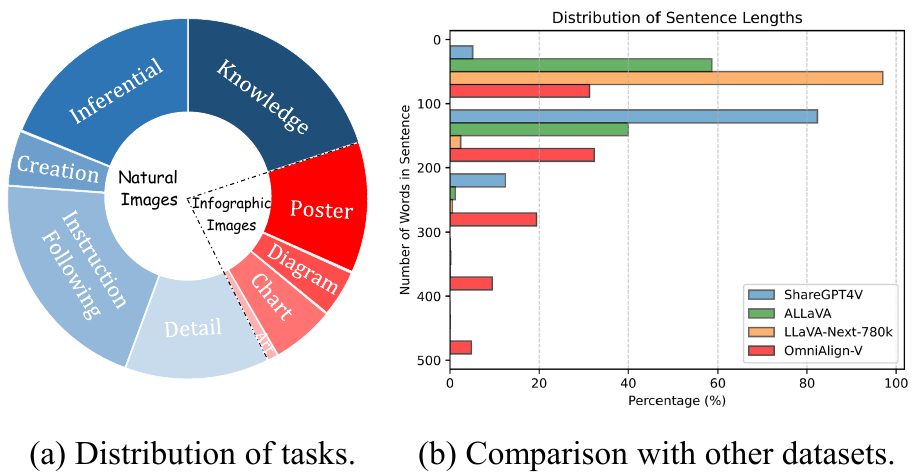}
    \caption{\textbf{Data distribution of OmniAlign-V.} Our dataset includes a diverse range of tasks, characterized by a more balanced distribution of answer lengths compared to those observed in ALLaVA and ShareGPT4V.}
    \label{fig:data_combination}
    \vspace{-10pt}
\end{figure}

Image content plays a crucial role in constructing multi-modal training data. 
To ensure comprehensive coverage, we classify images into two major categories: natural images and infographic images, as shown in \cref{fig:data_combination}(a). 
Our data synthesis pipeline first determines whether an input image belongs to natural images (captured from real-world scenes) or infographic images (human-created to convey information). 
Based on this classification, different vision-language tasks are assigned.

For natural images, we define three primary tasks: \textbf{Knowledge}, \textbf{Inferential}, and \textbf{Creation}, 
each requiring diverse and complex question formats with comprehensive, reasoned responses. 
These tasks enhance the model's ability to interpret real-world scenes effectively.

For infographic images, given their diverse content, 
we identify four key types that elicit intricate and challenging questions: \textbf{Arts}, \textbf{Charts}, \textbf{Diagrams}, and \textbf{Posters}. 
These categories necessitate a deep understanding of human-designed visuals, 
including both abstract and detailed elements.


\subsection{Image Selection Strategy}
\label{sec: filter}

For natural images, rich semantic content leads to more comprehensive and insightful QA pairs. 
To enhance training data quality, 
we developed a pipeline to select semantically rich images from diverse sources, 
as shown in \cref{fig:pipeline}(a). This pipeline consists of two key steps.

First, we use the Image Complexity (IC) recognition model IC9600~\cite{feng2022ic9600} to assign IC scores to images. 
Images with low semantic content —- 
characterized by few objects or simple, uniform backgrounds —- 
receive low IC scores and are then excluded. 
While IC9600 effectively filters out low-complexity images, high IC scores alone do not guarantee semantic richness. 
For instance, as shown in \cref{fig:imagefilter}, an image filled with tents may have a high IC score but lacks sufficient semantic information for multi-modal training.

To refine selection, we employ the Recognize Anything Model~\cite{zhang2024recognize} to identify objects within images, filtering out those with high complexity but minimal meaningful content. This two-step approach ensures our pipeline accurately selects images that are both complex and semantically rich. 
Experiment results have validated the effectiveness of our proposed image selection strategy, as shown in \cref{appx: filterablation}.

\subsection{Data Generation Pipeline}
\label{sec: sft generate}

\noindent\textbf{SFT QA Generation. } 
In \cref{fig:pipeline}(b), we outline the generation process of SFT QA pairs.
For vision-language tasks involving natural images (knowledge, creative, inferential), 
we first apply our image selection strategy to filter semantically rich images from CC-3M\cite{sharma2018conceptual}, Flickr30k \cite{plummer2015flickr30k}, and GQA~\cite{hudson2019gqa}.
For infographic image tasks, images are collected from various existing sources~\cite{masry2022chartqa,Li2018TextbookQA,kembhavi2016diagram}. 
More details are provided in \cref{sources}.


\textit{Knowledge \& Inferential Tasks. }
In preliminary experiments, we found that GPT-4o can generate diverse, content-relevant questions for the two tasks when provided with well-designed few-shot prompts. 
Consequently, we carefully designed a single prompt for each task category, incorporating comprehensive instructions and selected few-shot examples. 
These prompts were then used with GPT-4o across diverse images to generate QA pairs.

\textit{Creative Tasks. }We noticed that a single prompt cannot generate sufficiently diverse, 
content-relevant creative questions. 
Therefore, we developed a more sophisticated pipeline inspired by Condor~\cite{cao2025condor}. 
First, we created a set of seed creative questions:
\begin{equation}
    \mathcal{Q}_s =\{Q_1, Q_2, ... Q_N\}
\end{equation}
where each seed question corresponds to a distinct creative task. 
Since directly using all seed questions as few-shot examples leads to repetition and lack of diversity, 
we employ a light-weight MLLM~\cite{luo2024mono} to generate detailed captions $C$ for each image. An LLM $\mathcal{M}$ then selects a relevant subset of seed questions according to the caption:
\begin{equation}
    \mathcal{Q}_s' = \mathcal{M}(C, \mathcal{Q}_s), \|Q_s'\| \ll \|Q_s\|
\end{equation}
Finally, we randomly select three question types from $\mathcal{Q}_s'$ as few-shot examples for GPT-4o, preserving both quality and diversity of synthesized data.

 
\textit{Infographic Tasks. }
For infographic image tasks (\textit{Charts}, \textit{Diagrams}, \emph{etc.}), 
questions and answers are closely tied to specific image content. 
Unlike natural images, these visuals convey information primarily through text, colors, lines, and symbolic elements, 
making evaluation based on image complexity or object categories unsuitable. Therefore, instead of applying image selection strategies, 
we carefully select image sources containing rich, detailed information. 
We then design specialized prompts for GPT-4o to generate questions that require comprehensive background knowledge understanding. The difference is shown below:

\noindent\textit{InfographicVQA: What is the respiratory disease death rate for individuals aged 70+? }

\noindent\textit{OmniAlign-V: How does the respiratory disease death rate for individuals aged 70+ compare to the other age groups, and what might this suggest?}


\noindent\textbf{Post Refinement. } 
To further improve the quality of the synthesized data, 
 we implemented a series of post-processing methods for refinement,
as illustrated in \cref{fig:pipeline}(c).

\textit{Instruction Augmented Knowledge QAs. }
Instruction following is a crucial capability significantly impacting human preference. 
To enhance this, we incorporate complex instructions and restrictions into our knowledge QA pairs and reformulate responses accordingly. 
As shown in \cref{fig:pipeline}(c1), for each knowledge QA, 
we use a powerful LLM to select an appropriate instruction type that can be integrated into the question without depending on visual content. 
The instruction is then incorporated into the existing question, and an LLM adjusts the corresponding answer to ensure alignment with both the modified question and original context, resulting in instruction-augmented knowledge QAs.


\textit{Enriched Inferential QAs. }
For many inferential QAs, 
the answers lack sufficient detail to fully explain underlying logic and background knowledge. 
To address this, we employ a knowledge-rich LLM to enrich responses with detailed explanations, relevant background information, and logical reasoning (\cref{fig:pipeline}(c2)). 
This refinement enhances alignment with user preferences.


\textit{Quality Improved Infographic QAs. }
For infographic tasks, particularly those involving \textit{Chart} data, 
we observed that even SOTA MLLMs struggle with complex charts and detailed questions.
While GPT-4o excels at explaining background knowledge but often produces inaccurate OCR results, 
SOTA open-source MLLMs~\cite{chen2024far,bai2023qwen} show superior OCR accuracy but lack detailed explanations. 
Therefore, we developed a refinement pipeline to generate responses combining rich background knowledge and accurate OCR results. 
We filter out questions where GPT-4o and SOTA open-source MLLM responses show significant discrepancies in trends. 
For remaining questions, the responses from different MLLMs are merged to produce a final response that is both precise and comprehensive.
The merged answers are further reviewed by human experts to ensure their quality and consistency. More details are demonstrated in \cref{appd: chart post}.


OmniAlign-V comprises 39K Knowledge QAs, 37K Inferential QAs, 10K Creative QAs, 
38K Instruction-Following (Knowledge) QAs, and 44K Infographic QAs (2K Art, 8K Diagram, 11K Chart, 23K Poster). 
Additionally, we prompt GPT-4o to generate 35K QAs focusing on image details, resulting in a total of 205K high-quality SFT training samples. Examples are shown in \cref{fig:datasample1,fig:datasample2,fig:datasample3,fig:datasample4}.

\input{tables/main}
\input{tables/main_language}
\noindent\textbf{DPO Data Generation. } 
OmniAlign-V's high-quality, human-aligned QA pairs can serve as positive samples for DPO training. 
Inspired by Reject Sampling~\cite{casella2004generalized}, 
we generate negative samples by prompting a LLaVA-Next baseline (generator $G$) trained on LLaVA-Next-778k. 
For each question $Q_i$, the generator produces $N$ responses with high temperature,
$\mathbf{R} = \{r^i_1, r^i_2,... r^i_N\}$. An LLM Judger $J$ then evaluates these responses to select the one that most  deviates from the original question's intent and context as the negative sample $r^i_{Neg}$, ensuring clear contrast between positive and negative samples.



\begin{figure}[t]
    \centering
    \includegraphics[width=\linewidth]{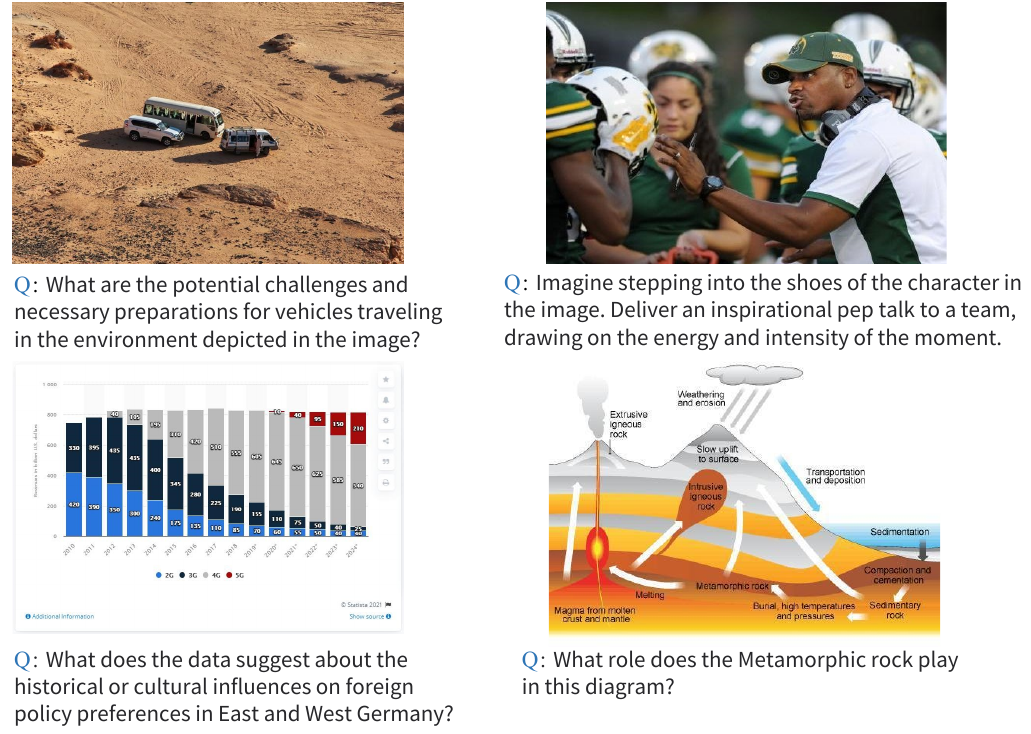}
    \caption{\textbf{Samples in MM-AlignBench}.}
    \label{fig:bench}
    \vspace{-10pt}
\end{figure}

\section{MM-AlignBench}

Current benchmarks for assessing multi-modal alignment capabilities are limited. While WildVision~\cite{lu2024wildvision} aims to evaluate human preferences in real-world interactions, 
it employs repetitive and simplistic question formats that inadequately assess response quality, as shown in \cref{fig:badbench}. 
To address this, we developed MM-AlignBench, featuring human-curated images and questions for more nuanced evaluation.

We selected high-quality images from SAM-1B~\cite{kirillov2023segment}, CC-3M-Test~\cite{sharma2018conceptual}, AI2D~\cite{kembhavi2016diagram}, ChartQA~\cite{masry2022chartqa}, and InfographicVQA~\cite{mathew2022infographicvqa}. 
For natural images, we applied our image selection strategy from \cref{sec: filter} to identify 2,000 semantically rich images and combine them with 1,000 carefully selected infographics images. 
GPT-4o was then used to generate diverse questions for these images. 
Human experts reviewed and refined the image-question pairs, filtering out low-quality, repetitive, or contextually weak samples. 
This process resulted in 252 high-quality question-image pairs, 
featuring diverse question types and semantically rich images.
Several examples are shown in \cref{fig:bench}.

For evaluation, we follow WildVision’s approach, using GPT-4o as the judge model to compare model responses with reference responses generated by Claude3V-Sonnet~\cite{Claude3}, reporting winning rates and reward scores.

%% file: tables/main.tex
\begin{table*}[t]
    \centering
    \resizebox{\textwidth}{!}{%
    \tablestyle{4pt}{1.1}
    \begin{tabular}{l|l|l|ccc|ccccc}
    \Xhline{0.15em}
        \textbf{Model} & \textbf{Data} & \textbf{LLM} &\textbf{MM-AlignBench}  &\textbf{WildVision} &\textbf{MIA-Bench} &\textbf{MMVet} & \textbf{MMMU} &\textbf{MMBenchV1.1}  &\textbf{AI2D} &\textbf{OCRBench} \\
        \hline
        LLaVA & LLaVANext-778k & InternLM2.5-7B         &3.6 / -82.1 &18.4 / -55.1 &75.4  &41.2 &42.6 &73.6 &74.1 &39.7 \\
        \rowcolor{gray!20}
        LLaVA & OmniAlign-V$_{mix}$ & InternLM2.5-7B    &50.0 / +3.8 &28.2 / -34.6 &85.4  &43.5 &43.3 &73.7 &74.7 &41.3  \\
        &                     &                  &\color{SeaGreen}{+ 46.4 / 85.9}&\color{SeaGreen}{+ 9.8 / 20.5}  &\color{SeaGreen}{+ 10.0}   &\color{SeaGreen}{+ 2.3} &\color{SeaGreen}{+ 0.7}&\color{SeaGreen}{+ 0.1}  &\color{SeaGreen}{+ 0.6} &\color{SeaGreen}{+ 1.6}  \\
        \hline
        LLaVANext & LLaVANext-778k & InternLM2.5-7B     &20.6 / -42.7 &23.4 / -45.0 &76.9  &41.8 &44.1 &75.1 &74.7 &56.2  \\
        \rowcolor{gray!20}
        LLaVANext& OmniAlign-V$_{mix}$ &InternLM2.5-7B  &57.1 / +11.1 &29.6 / -31.3 &86.7  &47.7 &46.8 &74.9 &77.5 &58.9  \\
        &                     &                  &\color{SeaGreen}{+ 36.5 / 53.8}&\color{SeaGreen}{+ 6.2 / 13.7}  &\color{SeaGreen}{+ 9.8}   &\color{SeaGreen}{+ 5.9} &\color{SeaGreen}{+ 2.7}&\color{SeaGreen}{- 0.2}  &\color{SeaGreen}{+ 2.8} &\color{SeaGreen}{+ 2.7}  \\
        \hline
        LLaVANext & LLaVANext-778k & Qwen2.5-32B        &26.6 / -29.0 &25.2 / -41.3 &86.0 &47.7 & 55.2 &79.3 &79.6 &55.9  \\
        \rowcolor{gray!20}
        LLaVANext & OmniAlign-V$_{mix}$   & Qwen2.5-32B &62.3 / +19.4 &40.2 / -14.9 &89.6  &56.9 & 60.7 &80.6 &81.7 &55.9  \\
        &                     &                  &\color{SeaGreen}{+ 35.7 / 48.4}&\color{SeaGreen}{+ 15.0/26.4}  &\color{SeaGreen}{+ 3.6}   &\color{SeaGreen}{+ 9.2} &\color{SeaGreen}{+ 5.5}&\color{SeaGreen}{+ 1.3}  &\color{SeaGreen}{+ 2.1} &\color{SeaGreen}{+ 0.0}  \\
    \Xhline{0.15em}
    \end{tabular}
    }%
    \caption{\textbf{Evaluation Results of Integrating OmniAlign-V into the SFT stage}. By integrating OmniAlign-V, the multi-modal preference alignment of MLLMs significantly improved. Additionally, we also observe comparable or better performance on common VQA benchmarks. In `Model' column, LLaVA and LLaVANeXT denote the model structure and training strategy. For MM-AlignBench and WildVision, notation \textit{A/B} denotes \textit{Winning Rate/Rewards}.}
    \label{tab: main}
    \vspace{-10pt}
\end{table*}

%% file: tables/main_language.tex
\begin{table}[t]\small
    \centering
    \resizebox{.5\textwidth}{!}{%
    \tablestyle{4pt}{1}
    \begin{tabular}{l|l|cc}
    \shline
        \multirow{2}{*}{\textbf{Model}} & \multirow{2}{*}{\textbf{Type}} &\textbf{AlpacaEvalv2}  &\textbf{ArenaHard}  \\
                       &   & \multicolumn{2}{c}{\textit{v.s. GPT-3.5/GPT-4}} \\
        \hline
        LLaVANext$^I$ & MLLM    & 29.8 / 3.8 & 21.4 / 4.93  \\
        \rowcolor{gray!20}
        LLaVANext$^I$-OA&MLLM &50.1 / 7.8 &30.4 / 9.33  \\
        InternLM2.5-7B  & LLM        &78.3 / 26.2 & 47.5 / 19.1   \\
        \hline
        LLaVANext$^Q$ & MLLM     &50.6 / 7.0 &54.5 / 18.0 \\
        \rowcolor{gray!20}
        LLaVANext$^Q$-OA& MLLM &77.6 / 18.0 &87.2 / 58.1 \\
        Qwen2.5-32B  & LLM        &92.1 / 37.1 &94.8 / 75.9 \\
    \shline
    \end{tabular}
    }%
    \caption{\textbf{Performance on text-only alignment benchmarks.} OA denotes models trained with OmniAlign-V.
    We present the winning rate against GPT-3.5 and GPT-4 (original setting).  OmniAlign-V can also enhance MLLM's performance on text-only alignment benchmarks.}
    \label{tab: main_lang}
    \vspace{-10pt}
\end{table}

%% file: latex/5_main_results.tex
\section{Evaluation Results}

\subsection{SFT with OmniAlign-V }

We conduct extensive experiments to demonstrate OmniAlign-V's effectiveness. 
We combine OmniAlign-V with LLaVA-Next-778k (excluding text-only samples), 
creating OmniAlign-V$_{mix}$ with 946K training samples. 
We evaluate various MLLMs tuned on OmniAlign-V against their counterparts tuned on LLaVA-Next-778k.

Our evaluation spans multiple multi-modal benchmarks, 
including standard VQA benchmarks~\cite{yu2023mm,liu2024mmbench,liu2023hidden,yue2023mmmu,kembhavi2016diagram} and human-preference alignment benchmarks: MM-AlignBench, WildVision ~\cite{lu2024wildvision}, and MIA-Bench~\cite{qian2024mia}. 
Results in \cref{tab: main} show that OmniAlign-V significantly improves human alignment across all benchmarks. 
Moreover, our training data improves general multi-modal capabilities, particularly on MMVet and MMMU, demonstrating a trend distinct from text-only data. 

Notably, despite excluding language samples from training data, models maintain stronger language alignment than those trained on LLaVA-Next-778k, as shown in Table~\ref{tab: main_lang}. 
This suggests that while high-quality language data alone may not significantly impact multi-modal capabilities, 
enhancing multi-modal data quality can improve both language and multi-modal performance, highlighting the crucial role of high-quality, human-aligned multi-modal training data.

\input{tables/dpo}

\subsection{DPO with OmniAlign-V-DPO}
We conduct DPO post-training on three models: 
LLaVA-Next trained with LLaVA-Next-778k, 
LLaVA-Next trained with OmniAlign-V$_{mix}$, 
and InternVL2-8B. 
Results in \cref{tab: dpo} show that DPO tuning significantly improves performance on real-world questions (WildVision) across all models. 
While the baseline trained solely on LLaVA-Next-778k shows minimal improvement on MM-AlignBench, 
models incorporating OmniAlign-V during SFT demonstrate substantial gains after DPO. 
Similarly, InternVL2-8B, a state-of-the-art MLLM partially trained on proprietary multi-modal corpora, shows significant improvement on MM-AlignBench post-DPO. 
This indicates that if a model has been trained on data aligned with human preferences, such as open-ended or long-context data  during SFT phase, the subsequent DPO training on high-quality human-aligned data can significantly activate the model’s ability, leading to a considerable improvement in alignment performance. 
In contrast, if the model has not been exposed to such alignment-focused datasets during SFT, training with open-ended data alone via DPO will not significantly improve its capabilities of alignment.
These findings demonstrate the value of OmniAlign-V in both SFT and DPO stages for enhancing human preference alignment.


\input{tables/benchmark_results}

\subsection{MM-AlignBench}
We evaluate various state-of-the-art MLLMs~\cite{2022chatgpt,team2023gemini,Claude3,bai2023qwen,chen2024far,li2024llava,minicpm2024,internlmxcomposer2,LLaMA32Vision,laurençon2024building,wang2024enhancing} on MM-Alignbench, with results shown in \cref{tab: bench}. 
Closed-source models like GPT, Claude, and Gemini demonstrate strong alignment in responding to open-ended questions. 
In contrast, while Qwen2-VL and InternVL2 excel at common VQA tasks, they show relatively lower human preference alignment. 
This highlights the importance of prioritizing MLLM alignment for improved daily human interactions. 
Our LLaVA-OA-32B, trained with OmniAlign-V, achieves exceptional performance, outperforming numerous strong MLLMs and nearly matching Qwen2VL-72B. 
After applying DPO with OmniAlign-V-DPO, \textbf{LLaVA-OA-32B-DPO achieves winning rate of 72.6 with an average reward of +33.5,  surpassing the performance of Qwen2VL-72B}. 
These results highlight the high quality and effectiveness of the OmniAlign-V dataset in improving model alignment with human preferences.


\subsection{Ablation Study}
We conduct an ablation study to evaluate each subset of OmniAlign-V, 
reporting results on MM-Alignbench, WildVision, and MMVet in \cref{tab: ablation}. 
Performance improves progressively as different tasks from OmniAlign-V are incorporated. 
Notably, Instruction-Following data significantly enhances performance across all three benchmarks, 
demonstrating its crucial role. 
The creation data subset uniquely improves performance on MM-Alignbench but not on WildVision and MMVet.
This discrepancy can be attributed to the absence of multi-modal creative question types in these two benchmarks, suggesting their incompleteness in capturing full spectrum of alignment challenges.
\input{tables/ablation_data}


\section{Conclusion}
In this paper, we introduce \textbf{OmniAlign-V}, 
a dataset designed to enhance the alignment of MLLMs with human preferences, 
as well as \textbf{MM-AlignBench}, 
a high-quality, specific-purpose benchmark for evaluating such alignment. 
We investigate the impact of both language and multi-modal training data, 
emphasizing the critical role of multi-modal open-ended training data. 
By incorporating OmniAlign-V into SFT and DPO stages, we achieve significant improvements in the alignment of MLLMs. 

\section{Limitation}
Although the OmniAlign-V pipeline can be easily scaled to support larger datasets, the scale of the dataset used in this paper may be insufficient for large-scale training due to cost limitations. 
Deeper exploration into the alignment of MLLMs is still needed to address these limitations and further advance the field.

%% file: tables/dpo.tex
\begin{table}[t]
    \centering
    \resizebox{.48\textwidth}{!}{%
    \tablestyle{4pt}{1.2}
    \begin{tabular}{l|l|ccc|ccccc}
    \Xhline{0.11em}
        \textbf{Model} & \textbf{Stage} &\textbf{MM-AlignBench}  &\textbf{WildVision}\\
        \hline
        LLaVANext$^I$ & SFT            &9.5 / -69.2 &30.4 / -34.2 \\
        \rowcolor{gray!20}
        LLaVANext$^I$ & SFT+DPO        &11.1 / -64.5 & 35.5 / -23.4 \\
        \hline
        LLaVANext$^I$-OA & SFT       &57.1 / +11.1 &29.6 / -31.3   \\
        \rowcolor{gray!20}
        LLaVANext$^I$-OA & SFT+DPO   &64.3 / +22.4 &41.8 / -10.1  \\
        \hline
        InternVL2-8B   & SFT                &31.4 / -21.8 & 48.6 / +1.4\\
        \rowcolor{gray!20}
        InternVL2-8B   & SFT+DPO            &64.7 / +19.4 & 51.4 / +1.9\\
    \Xhline{0.11em}
    \end{tabular}
    }%
    \caption{\textbf{Performance of applying DPO with OmniAlign-V-DPO}. For models finetuned with human-aligned data, by employing DPO training, the model's alignment with human perference further improved.}
    \label{tab: dpo}
    \vspace{-10pt}
\end{table}

%% file: tables/benchmark_results.tex
\begin{table}[t]
    \centering
    \resizebox{.5\textwidth}{!}{%
    \tablestyle{3pt}{1.1}
    \begin{tabular}{l|cc|ccccc}
    \Xhline{0.15em}
        \textbf{Model} &\textbf{Win Rate} $\uparrow$ &\textbf{Reward}$\uparrow$ &B+ & B& T& W& W+\\
        \hline
        Claude3.5V-Sonnet & 84.9 & +51.4 & 70 & 144 & 12 & 31 & 4 \\ 
        GPT-4o &81.3 &+49.0 & 81 & 124 & 12 & 31 & 4\\
        GPT-4V &82.5 &+46.0 & 57 & 157 & 12 & 31 & 1\\
        GeminiFlash1.5-002 &77.0 &+39.1 & 56 & 138 & 14 & 35 & 9\\
        \rowcolor{gray!20}
        \textbf{LLaVANext-OA-32B-DPO} & 74.2 & +36.9 & 49 & 138 & 20 & 40 & 5 \\
        Qwen2VL-72B   &61.5 &+21.6 & 43 & 112 & 15 & 75 & 7\\
        \rowcolor{gray!20}
        \textbf{LLaVANext-OA-32B} & 62.3 & +19.4 & 31 & 126 & 19 & 62 & 14 \\
        \hdashline
        Claude-3V-Sonnet & 50 & 0  & - & - & - & - & -  \\
        \hdashline
        Qwen2VL-7B   &44.4 & -5.8 & 28 & 84 & 5 & 101 & 34\\
        InternVL2-72B &44.4 &-6.9 & 19 & 93 & 8 & 98 & 34\\
        InternVL2-8B-MPO &40.1 & -10.9 & 26 & 75 & 10 & 100 & 41\\
        InternVL2-8B &31.3 & -21.8 & 18 & 61 & 15 & 109 & 49\\
        LLaMA3.2-Vision-11B & 27.8&-33.7 & 18 & 52 & 4 & 98 & 80\\
        \rowcolor{gray!20}
        \textbf{LLaVANext-Qwen32B} & 26.6 & -29.0 & 16 & 51 & 10 & 121 & 54 \\
        LLaVA-OneVision-7B &23.8 &-46.2 & 14 & 46 & 1 & 75 & 116 \\
        MiniCPM-V-2.5 &12.7 & -53.0 & 9 & 23 & 8 & 116 & 96 \\
        Xcomposer2.5-7B &7.5 & -74.0 & 5 & 14 & 3 & 63 & 167\\
        Idefics3-8B     &2.7 & -92.3 & 3 & 4 & 0 & 15 & 230\\
        \Xhline{0.15em}
    \end{tabular}
    }%
    \caption{\textbf{Performance of existing MLLMs on MM-AlignBench}. B+/B/T/W/W+ denotes MuchBetter/Better/Tie/Worse/MuchWorse.  Our LLaVA-Next-OmniAlign(OA)-32B-DPO, trained with OmniAlign-V and applied DPO with OmniAlign-V-DPO, demonstrates outstanding performance, surpassing a wide range of strong MLLMs, even Qwen2VL-72B.}
    \label{tab: bench}
    \vspace{-10pt}
\end{table}

%% file: tables/ablation_data.tex


\begin{table}[t]
    \centering
    \resizebox{.5\textwidth}{!}{%
    \begin{tabular}{l|ccc}
    \Xhline{0.12em}
        \textbf{Model}  &\textbf{MM-AlignBench}  &\textbf{WildVision} &\textbf{MMVet}\\
        \hline
        LLaVANext-778k                   & 20.6 / -42.7 & 23.4 / -45.0 & 41.7\\
        +Know / Infer / Detail               & 23.4 / -42.1 & 23.2 / -43.2 & 42.8\\
        +Instruction Following           & 36.5 / -17.3 & 26.2 / -37.5 & 44.6\\
        +Creation                        & 43.7 / -5.0 & 26.6 / -37.7 & 44.4 \\
        +Chart / Diagram / Poster            & 57.1 / +11.1 & 29.6 / -31.3 & 47.7\\
    \Xhline{0.12em}
    \end{tabular}
    }%
    \caption{\textbf{Ablation of Subsets in OmniAlign-V.} By progressively incorporating tasks within OmniAlign-V, the model's alignment performance gradually improves.}
    \label{tab: ablation}
    \vspace{-10pt}
\end{table}

%% file: latex/6_appd.tex
\newpage
\newpage
\newpage
\appendix

\section{Alignment of MLLMs with Human Preference}
\label{appd: align}
\begin{figure*}[t]
    \centering
    \includegraphics[width=\linewidth]{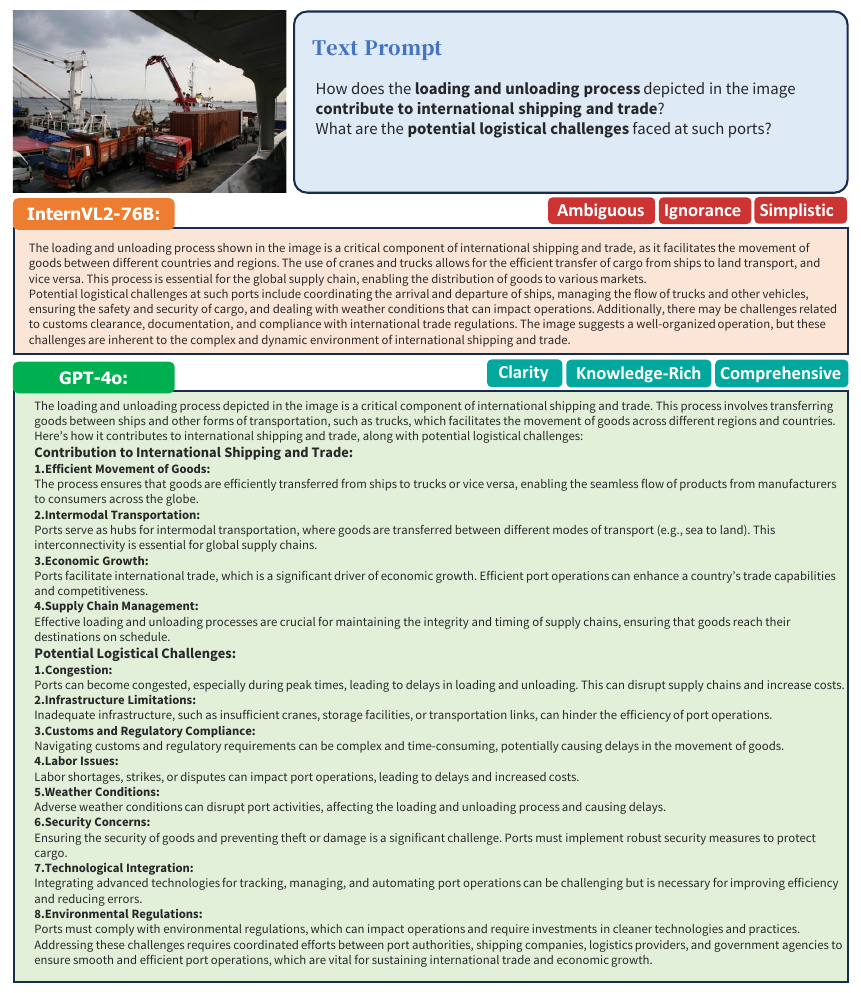}
    \caption{GPT-4o shows superior alignment with human preference than InternVL2-76B.}
    \label{fig:internvl}
\end{figure*}

Although open-source MLLMs have already matched or even surpassed proprietary models like the GPT and Claude series in common VQA tasks like OCR and Visual Perception, a significant gap remains in their alignment with human preferences. When posed with open-ended questions that require knowledge-rich responses, even the most advanced open-source MLLM, InternVL2-76B, struggles to provide comprehensive answers with high readability, as illustrated in Fig. \ref{fig:internvl}. In contrast, GPT-4o not only accurately identifies the main objects relevant to the question but also provides well-structured responses enriched with comprehensive contextual knowledge, achieving a high level of alignment with human preferences.

\section{Image sources}
\label{sources}
We carefully select image sources for \textit{Arts}, \textit{Charts}, \textit{Diagrams}, and \textit{Posters} tasks.
For \textit{Arts}, WikiArt~\cite{wikiart} is selected as the image source, offering a diverse range of painting styles. We uniformly sample 2000 images across all painting styles to ensure diversity in the dataset.
For \textit{Charts}, we select ChartQA~\cite{masry2022chartqa}, a dataset featuring charts that contain substantial statistical information. ChartQA includes several subcategories, from which we filter out simplistic charts with only two columns and retain those that contain charts with rich contextual information and diverse types.
For \textit{Diagrams}, we choose images from TextbookVQA~\cite{wikiart}, which provides diagrams rich in natural content and detailed explanations. We exclusively utilize the question images and teaching images from the image sources, as they meet our specific requirements.
For \textit{Posters}, we utilize InfographicsVQA~\cite{mathew2022infographicvqa}, a dataset containing high-quality posters with intricate designs and informative content.  


\section{Chart Post-Refine Pipeline}
\label{appd: chart post}
After curating images along with high-quality question, we first employ the current most powerful models(GPT-4o, InternVL2-72B, Qwen2VL-72B) to generate answers separately. 
A powerful LLM(Qwen2.5-72B) is then utilized to extract objective facts from the chart within each generated answer. 
The facts extracted from the answers of multiple models are compared for consistency. If the facts differ significantly and lead to entirely different conclusions, such responses are flagged for further review or discarded to avoid misinformation.
For cases where the facts exhibit only minor differences, we merge the detailed factual content from Qwen2VL-72B into the comprehensive explanations provided by GPT-4o. 
The merged answers are further reviewed and supervised by human experts to ensure their quality and consistency. 
\begin{figure}[t]
    \centering
    \includegraphics[width=\linewidth]{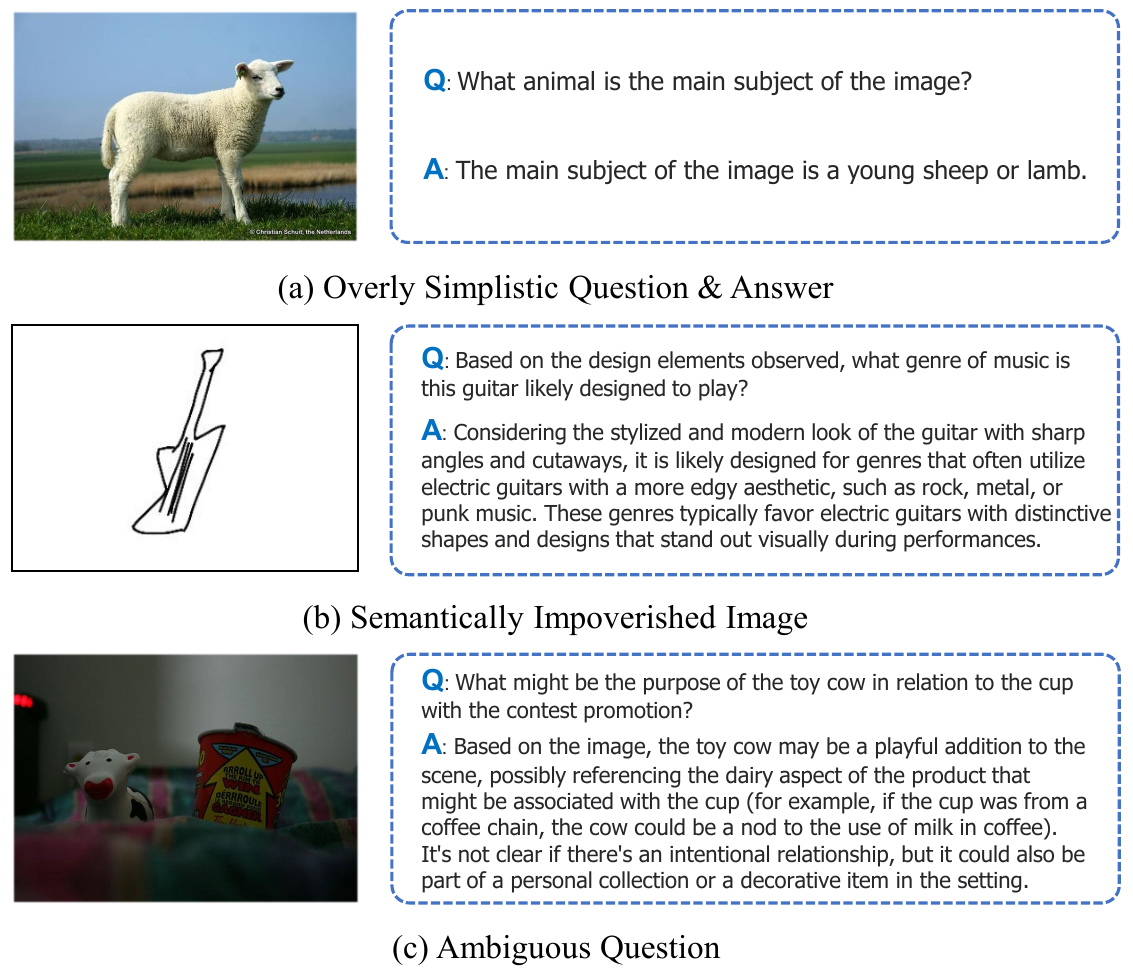}
    \caption{Examples of limitation with current multi-modal instruction tuning dataset.}
    \label{fig:badsft}
\end{figure}
\section{Training Details}
Our training strategy largely follows the approaches adopted by LLaVA and LLaVANext. CLIP-Large-336-Patch14 is employed as the visual encoder. In line with the LLaVA training strategy, we first conduct a pretraining stage where both the visual encoder and the LLM are frozen. 
We utilize the LLaVA-pretrain-558k  and ALLaVA-pretrain-728k  datasets for pretraining. The batch size is uniformly set to 256 and learning rate is set to 1e-3.
During this phase, images are resized to 336×336, and no image-splitting method is applied. 

For SFT stage, we unfreeze the LLM for LLaVA and further unfreeze the visual encoder for LLaVANext. 
In the case of LLaVANext, we apply the image-splitting method, setting the maximum split size to 3×3. The batch size is uniformly set to 128 and learning rate is set to 2e-5.. LLaVA-InternLM2.5-7B  is trained using 8×A800 GPUs for 12 hours. LLaVANext-InternLM2.5-7B  is trained using 16×A800 GPUs for 13 hours. LLaVANext-Qwen2.5-32B  is trained using 32×A800 GPUs for 24 hours.

\input{tables/ablation_image}
\section{Ablation on Image Filter}
\label{appx: filterablation}
We conduct ablation studies to evaluate the impact of using our image filter. We randomly sample 77k images from the LLaVA-Next-778K SFT dataset as a baseline and separately select images from CC3M both with and without applying our image filter. 
Subsequently, the selected images are used to generate Knowledge and Inferential Question-Answer pairs following the pipeline described in \cref{sec: sft generate}.  
We then assess the effect of these datasets on the performance of LLaVA-Next-InternLM2.5-7B, with the results presented in \cref{tab: ablation image}. 
It can be observed that by utilizing the image filter, the selected images contain richer semantic information, leading to the generation of higher-quality image-question pairs. This, in turn, enhances the model's performance in terms of alignment with human preferences.

\section{Human Experts}
In this study, two authors (both PhD students in computer science) serve as human experts. They are responsible for reviewing and refining the questions in MM-AlignBench, as well as evaluating and filtering incorrect merged cases following the Chart Post-Processing Refinement stage.

\section{License}
The InternLM and Qwen  models are licensed under the Apache-2.0  license. The ChartQA dataset is distributed under the GNU General Public License v3.0. The remaining datasets are licensed under CC BY-NC 4.0, which permits only non-commercial use.

\begin{figure*}[t]
    \centering
    \includegraphics[width=\linewidth]{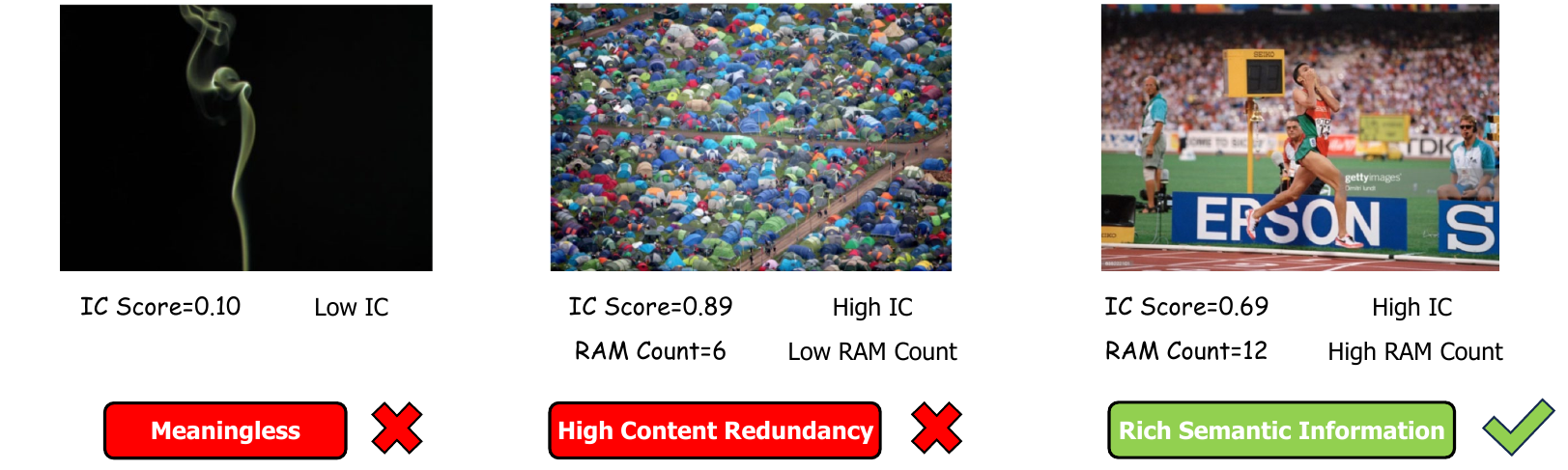}
    \caption{Demonstration examples of our image filter.}
    \label{fig:imagefilter}
\end{figure*}
\begin{figure*}[t]
    \centering
    \includegraphics[width=\linewidth]{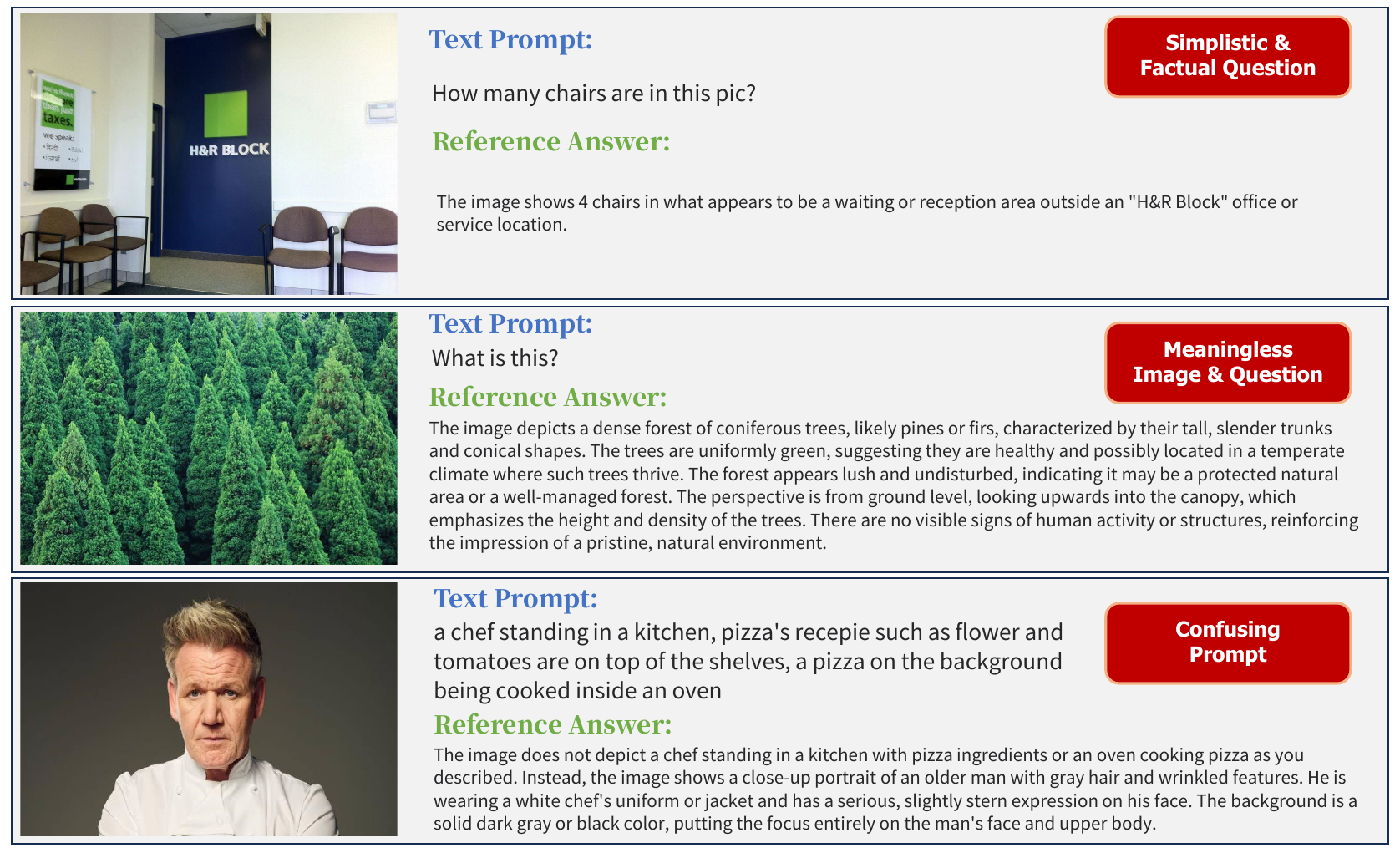}
    \caption{Examples of limitation within current multi-modal benchmark for alignment.}
    \label{fig:badbench}
\end{figure*}
\begin{figure*}[t]
    \centering
    \includegraphics[width=\linewidth]{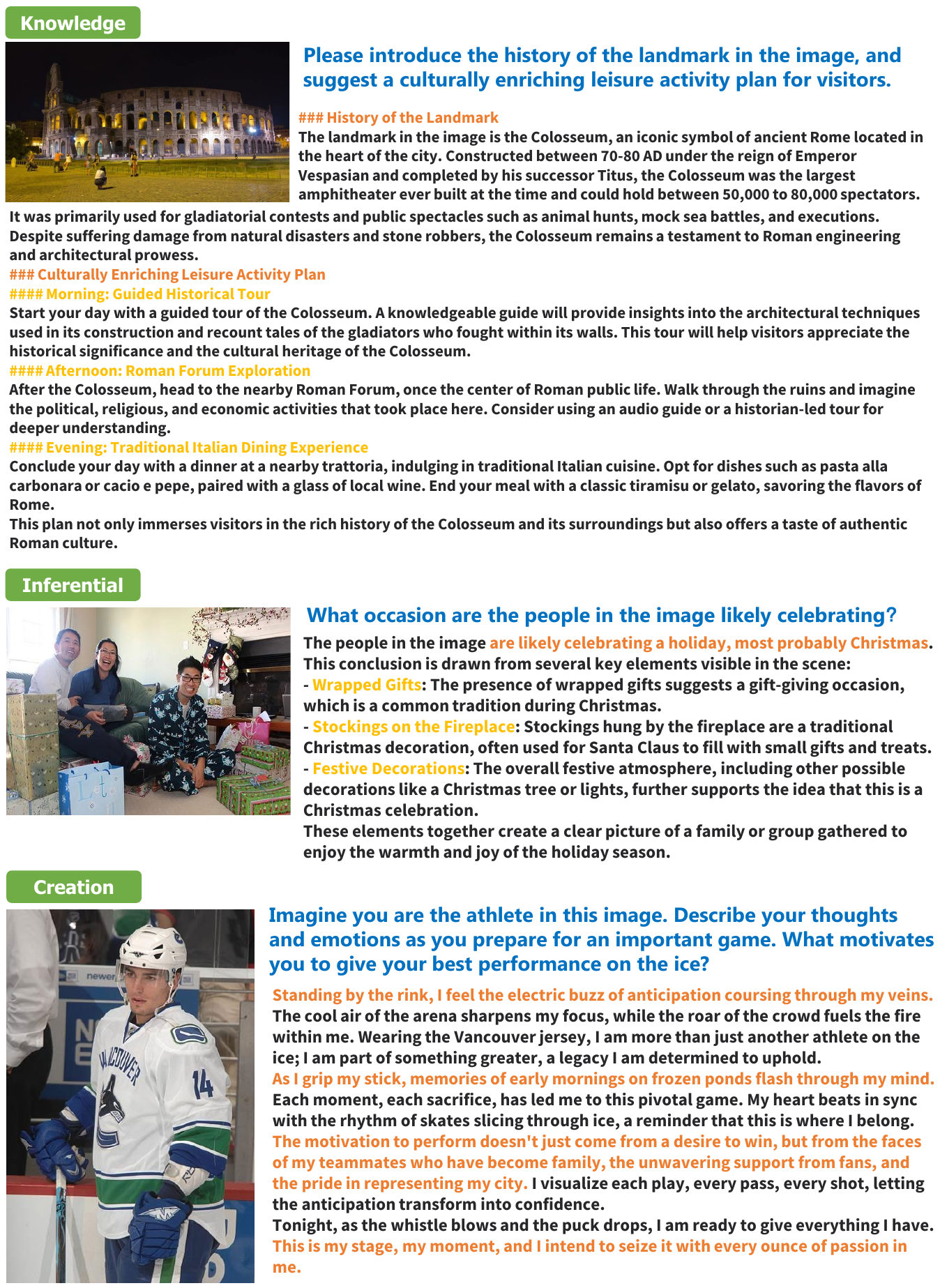}
    \caption{Examples of each task in OmniAlign-V.}
    \label{fig:datasample1}
\end{figure*}
\begin{figure*}[t]
    \centering
    \includegraphics[width=\linewidth]{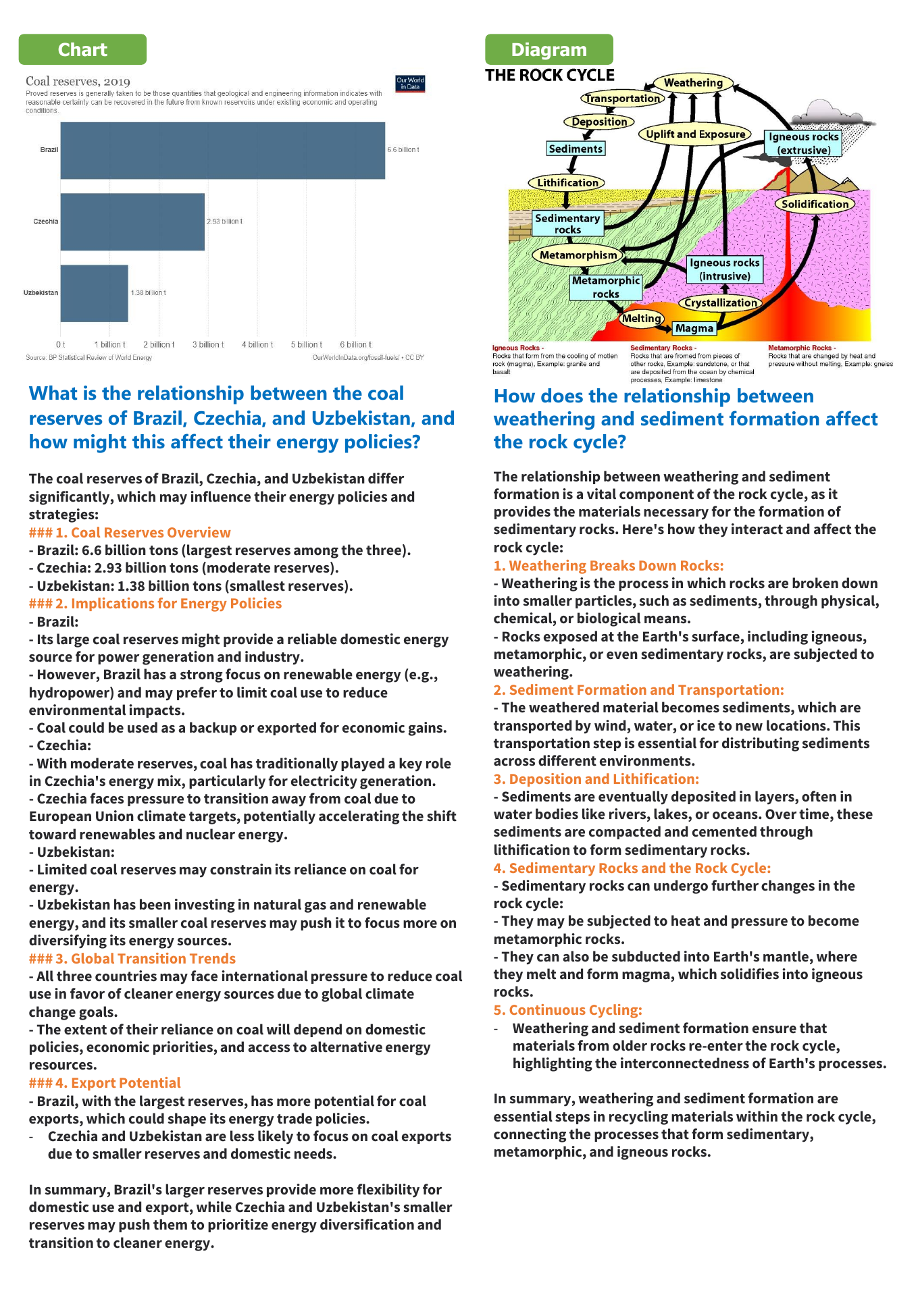}
    \caption{Examples of each task in OmniAlign-V.}
    \label{fig:datasample2}
\end{figure*}
\begin{figure*}[t]
    \centering
    \includegraphics[width=\linewidth]{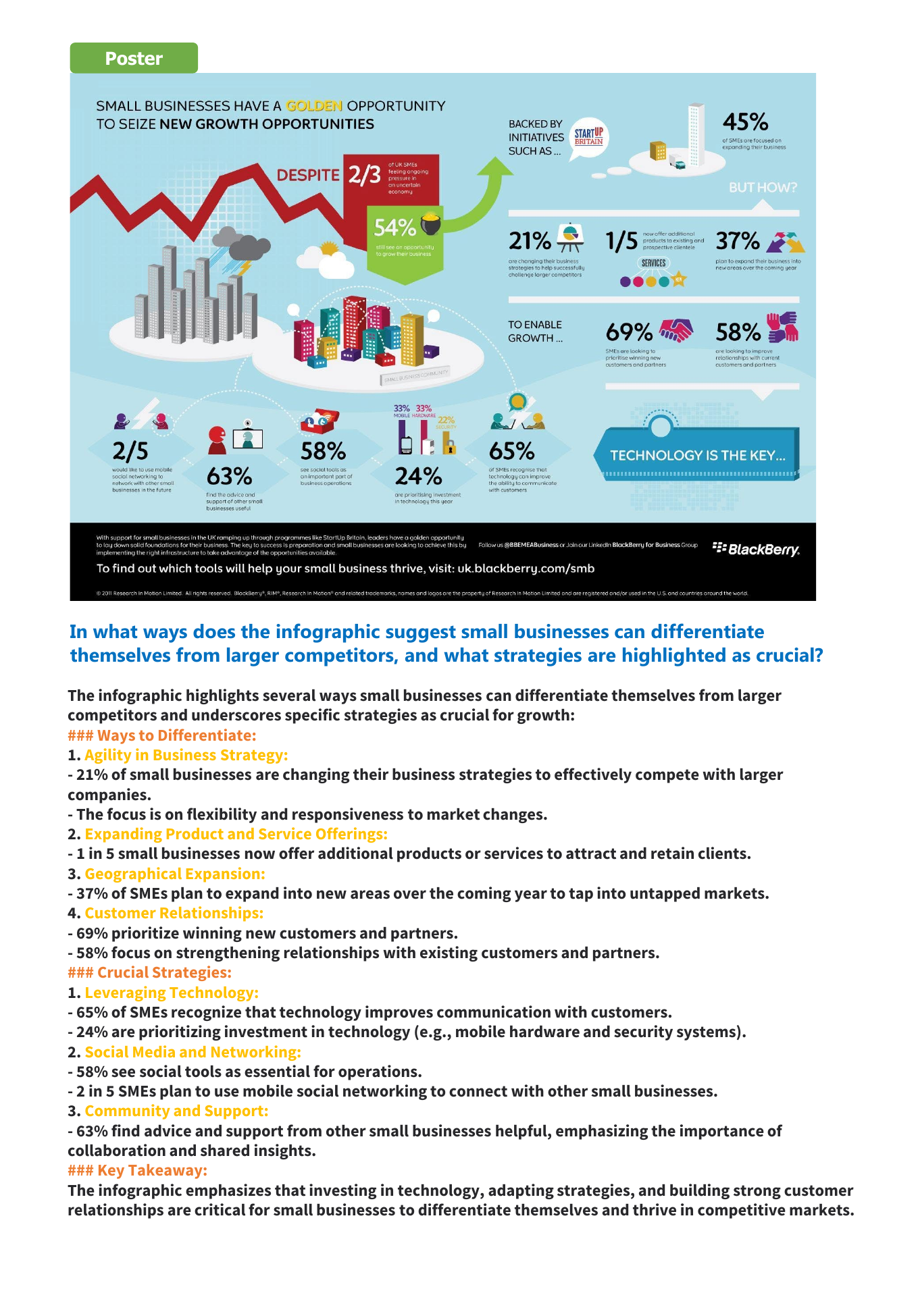}
    \caption{Examples of each task in OmniAlign-V.}
    \label{fig:datasample3}
\end{figure*}
\begin{figure*}[t]
    \centering
    \includegraphics[width=\linewidth]{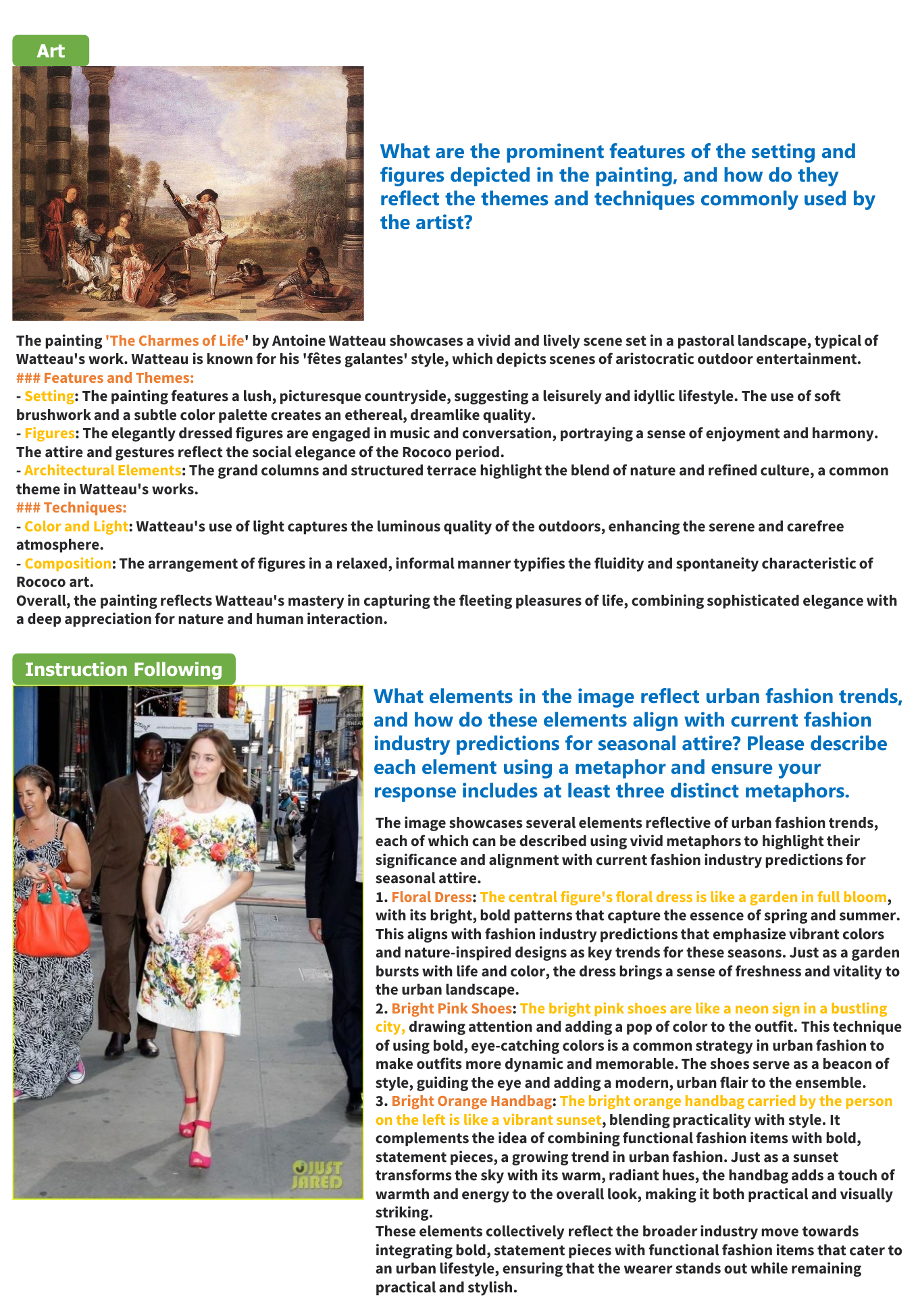}
    \caption{Examples of each task in OmniAlign-V.}
    \label{fig:datasample4}
\end{figure*}

\begin{figure*}[!ht] 
\begin{AIbox}{Prompt for Knowledge Task}
{Examine the image provided and generate a knowledge-based and exploratory question based on the content of the image and supply corresponding detailed answers.\\
 Question Guidelines:\\
- Your question should invite insightful discussion on the types of elements in the image, such as:\\
  - **Objects**: For example, animals, plants, food, or products.\\
    - E.g., "What breed of dog is in the picture, and what are their characteristics?", "Can you give me a recipe for the food in image?", "Write a Product Description for the product in the image.".\\
  - **Locations and Features**: Relevant to countries, landmarks, famous people, or scenic spots.\\
    - E.g., "Please introduce the history of the landmark in the picture.", "How did the states in image get their names?", "Who is the person in the image and what is him famous for?".\\
  - **Activities and Technologies**: Related to sports, machines, technology, and environmental details.\\
    - E.g., "Can you explain how the game in image is played?", "How is the machine shown in the image operated?"\\
  - **Events, Literature, and Media**: Concerning books, movies, or series in the image.\\
    - E.g., "Write a short description about the movie or series in the image.", "Think of books that would be enjoyable for someone who liked the books in the image."\\

 Answer Guidelines:\\
- Ensure your answers are factual and comprehensive.\\
- Please use Markdown formatting in your text to enhance the content, making it visually appealing and easy to read. Include appropriate headings, subheadings, lists, code blocks, and other Markdown elements to optimize your answers.\\
 Output Format:\\
Your response should strictly follow this format:\\
```json\\
\{\\
  "question": "Question text",\\
  "answer": "Answer text"\\
\}\\
```
}
\end{AIbox} 
\caption{\textbf{An Example of the prompt for Knowledge task generation. }}
\label{fig: prompt_kn}
\end{figure*}

\begin{figure*}[!ht] 
\begin{AIbox}{Prompt for Creative tasks}
{You are a skilled writer with a talent for crafting insightful and engaging questions based on the content of a given image.\\
**Task Guidelines**:\\
- Analyze the content of the provided image and select one of the question types listed below. Use it to create engaging questions that lead the viewer to explore and interpret the image in different level of complexity.\\
    - Be closely tied to the content of the image, emphasizing its primary visual or thematic elements.\\
    - Your questions should **avoid directly referencing specific details** in the image. Instead, they should encourage deeper reflection, ensuring the question cannot be answered without seeing the image.\\
    - Avoid overly rigid or direct phrasing, focusing instead on open-ended exploration.\\
**Question Types**\\
Your questions can be from the following types, each followed by an example with a different level of complexity.\\
- Simple (basic observation or initial reflection)\\
- Moderate (more thought-provoking, requiring a deeper understanding and more structured response)\\
- Difficult (complex or abstract, requiring analysis and strict formatting) \\
\textcolor{Red}{\{Match Types\}} \\
**Output Format**:\\
Your response should strictly follow this format:\\
```json\\
\{\\
  "question": "Question text",\\
  "type": "Question type",\\
  "level": "Question level"\\
\}
```
}
\end{AIbox} 
\caption{\textbf{An Example of the prompt for Creative task generation. }}
\label{fig: prompt_creation}
\end{figure*}

\begin{figure*}[!ht] 
\begin{AIbox}{Prompt for Inferential tasks}
{You are an image analysis expert skilled in posing high-quality **inferential questions**. Please provide 2-5 of the most insightful questions you can think of, following these guidelines:\\
For Questions:\\
- **Focus on image-based questions:** Ensure that your question cannot be answered without analyzing the image. **You should not directly provide image's data or details in your generated questions.** For example, "What might be the impact on the radio industry due to 34 stations having their licenses revoked?" includes the data in the image and can be answered without analyzing the image, so it is a bad question.\\
- Ensure that questions are natural, not overly rigid. **You should be quite certain and confident about the questions you pose and their answers, and avoid using words like "possibly", "maybe" or "might be" in both questions and answers.**\\
- Your questions must require reasoning beyond the direct content of the image, making reasonable inferences based on the information presented.
- The scope of the questions should not be overly broad or delve into political, philosophical, speculative, sensitive, or controversial topics. Stay within the context of the scene and elements inferred from it.\\
 For Answers:\\
- You should provide a clear and concise answer to the question.\\
Good Examples:\\
- What precautions are the people on the boat taking to stay comfortable during the trip?
- Is there anything else on the table other than the pizza?
- Why do these people choose to dress in this style?
- What decorative element is present in this public restroom that is not typical?
Bad Examples:\\
- What might indicate \\
}
\end{AIbox} 
\caption{\textbf{An Example of the prompt for Inferential task generation. }}
\label{fig: prompt_infer}
\end{figure*}

\begin{figure*}[!ht] 
\begin{AIbox}{Prompt for Chart tasks}
{You're a great image analyst. You need to analyze the image provided and generate some insightful questions based on the content of the image. \\
Question generation guidelines:\\
- Ensure that your questions require the image to be answered and do not include explicit information from the image. Instead, pose questions that prompt the respondent to analyze the image to find the answer.\\
- Your question should be explainable and require some reasoning to answer.\\
- Your question could contain different analytical perspectives, such as trends, comparisons, causal inference, etc.\\
- - Your questions should be insightful but also clear and straightforward. Avoid overly complex or niche questions.\\
Bad question examples:\\
- "What might be some factors contributing to the significantly higher private health expenditure per person in Argentina compared to Fiji and Benin?" (This includes specific details from the chart.) Its correct clarification should be "Is there any difference in private health expenditure per person between Argentina, Fiji, and Benin? If so, what might cause the difference?"\\
- "What trends can be observed in private health expenditures per person among the three countries shown?" (This question is unclear because 'trends between countries' is not a standard analytical concept. Trends typically refer to patterns over time or categories, not direct cross-entity comparisons.)\\
Output format:\\
Your response should strictly follow this format:\\
\{\\
"questions": [\\
\{\\
 "question": "Question text"\\
\},\\
\{\\
"question": "Another question text"\\
\}\\
]\\
\}\\
}
\end{AIbox} 
\caption{\textbf{An Example of the prompt for Chart task generation. }}
\label{fig: prompt_chart}
\end{figure*}

\begin{figure*}[!ht] 
\begin{AIbox}{Prompt for Poster tasks}
{You're an excellent image analyst and good at generating insightful questions about the image. You need to analyze the image provided and generate some insightful questions based on the content of the image, and you should answer the questions you generated.\\
Possible Categories Reference:\\
1. **Cultural and Social Context**\\
2. **Analysis and Inference**.\\
3. **Visual Elements and Design Techniques**.\\
Question Guidelines:\\
- **Focus on image-based questions:** Ensure that your question cannot be answered without analyzing the image. **You should not directly provide image's data or details in your generated questions.** For example, "What might be the impact on the radio industry due to 34 stations having their licenses revoked?" includes the data in the image and can be answered without analyzing the image, so it is a bad question.\\
- **Encourage thoughtful, structured responses:** Your question should be explainable and need some reasoning to answer. You should not generate questions that are just extracting information from the image. For example, "What percentage of organizations verify the past employment records of new employees according to the image?" is a bad question.\\
- **Ensure diversity in the questions:** Cover different aspects of the image, encouraging multiple perspectives. You can choose some appropriate categories from the possible categories reference. **For the same category, you can generate multiple questions.**\\
- **Generate high-quality questions:** You can choose to generate challenging questions, but their answers should be able to clearly explain. \\

Output Format:\\
Your response should strictly follow this format:\\

"questions": [\\
\{\\
"question": "Question text"\\
\}\\
]
}
\end{AIbox} 
\caption{\textbf{An Example of the prompt for Diagram task generation. }}
\label{fig: prompt_info}
\end{figure*}

\begin{figure*}[!ht] 
\begin{AIbox}{Prompt for Diagram tasks}
{You're a great diagram analyst. You need to analyze the diagram provided and generate 2-4 insightful questions based on the content of the diagram.\\
Question Guidelines:\\
- Your questions should be guiding and should not directly point to the content. For example: "How does acid rain affect water bodies, soil, and plant life?" should be changed to "How does the process in image affect water bodies, soil, and plant life?"\\
- Your question should invite insightful discussion, such as:\\
  - **Interpretation**: Symbol Interpretation, Data Extraction\\
    - **Examples**:\\
      - "What is the role of the cytokine-producing cell in the process shown?"\\
      - "Enumerate the steps outlined in the flowchart."\\
  - **Relation Analysis**:\\
    - **Examples**:\\
      - "How does variable A affect variable B in the diagram?"\\
      - "How many ways can A to B be achieved in the diagram?"\\
  - **Inference**:\\
    - **Examples**:\\
      - "What can be inferred about the system's stability from the diagram?"\\
      - "What does the bacterium do once it has the hybrid plasmid?"\\
Output Format:\\
Your response should strictly follow this format:\\
```json\\
\{\\
  "question":["Question text 1", "Question text 2", ...],\\
\}\\
```
}
\end{AIbox} 
\caption{\textbf{An Example of the prompt for Diagram task generation. }}
\label{fig: prompt_diagram}
\end{figure*}

%% file: tables/ablation_image.tex
\begin{table}[t]
    \centering
    \resizebox{.5\textwidth}{!}{%
    \begin{tabular}{l|ccc}
    \Xhline{0.15em}
        \textbf{Model}  &\textbf{MM-AlignBench}  &\textbf{WildVision} &\textbf{MMVet}\\
        \hline
        LLaVANext-77k                   & 15.1/-52.6  & 13.6 / -63.1 & 37.7\\
        + 33k w.o. Imager Filter              & 31.4 / -42.3 & 22.0 / -42.3 & 42.0\\
        + 33k w. Imager Filter            & 35.3 / -41.0 & 22.6 / -37.5 & 44.4\\
    \Xhline{0.15em}
    \end{tabular}
    }%
    \caption{Ablation study on the impact of utilizing image filter.}
    \label{tab: ablation image}
    \vspace{-10pt}
\end{table}